\pdfoutput=1

\documentclass[11pt]{article}

\usepackage[final]{acl}

\usepackage{times}
\usepackage{latexsym}
\usepackage{tabularx}
\usepackage{booktabs} 
\usepackage[T1]{fontenc}

\usepackage[utf8]{inputenc}

\usepackage{microtype}
\usepackage{pgf-pie}
\usepackage{tikz}
\usepackage{inconsolata}

\usepackage{graphicx}

\title{What Should Baby Models Read? Exploring Sample-Efficient Data Composition on Model Performance}

\author{Hong Meng Yam \\
  Stanford University \\
  \texttt{hongmeng@stanford.edu} \\\And
  Nathan Paek \\
  Stanford University \\
  \texttt{nathanjp@stanford.edu} \\}

\begin{document}
\maketitle
\begin{abstract}
 We explore the impact of pre-training data composition on the performance of small language models in a sample-efficient setting. Using datasets limited to 10 million words, we evaluate several dataset sources—including child-directed speech (CHILDES), classic books (Gutenberg), synthetic data (TinyStories), and a mix of these (Mix)—across different model sizes ranging from 18 million to 705 million parameters. Our experiments show that smaller models (e.g., GPT2-18M and GPT2-44M) benefit from training on diverse datasets like Mix, achieving better performance on linguistic benchmarks. In contrast, larger models (e.g., GPT2-97M, GPT2-705M, and LLaMA-360M) perform better when trained on more complex and rich datasets like Gutenberg. Models trained on the CHILDES and TinyStories datasets underperformed across all model sizes. These findings suggest that the optimal dataset for sample efficient training depends on the model size, and that neither child-directed speech nor simplified stories are optimal for language models of all sizes. We highlight the importance of considering both dataset composition and model capacity for effective sample efficient language model training. 
\end{abstract}

\section{Introduction}

In recent years, advancements in natural language processing have been largely driven by scaling language models to unprecedented sizes. Various large-language model (LLM) scaling laws have been formulated \cite{sardana2024chinchillaoptimalaccountinginferencelanguage}, with perhaps the most influential being the Chinchilla law, which demonstrates that parameters and tokens scale approximately linearly as the model scales \cite{hoffmann2022trainingcomputeoptimallargelanguage}. Many subsequent LLMs have been trained following this model \cite{rae2021scaling}, with some models including the Llama 2 and Llama 3 family of models being trained on 2 and 15 trillion tokens respectively, far more than the 'optimal' amount according to the Chinchilla scaling law \cite{dubey2024llama}. However, it is often prohibitive to train such large models, and impractical to continue scaling with the amounts of data required to train such models. 

This has sparked interest in small language models \cite{schick2020s, magister2022teaching} with much fewer parameters, requiring much less data for training. While much research has been conducted on knowledge distillation and improving the model architecture for small language models, comparably less research has investigated the contributions of different types of data used for model training, which is arguably just as important. Indeed, because LLM pretraining data typically comprises a mix of sources \cite{chowdhery2023palm}, researchers have found that the composition of pretrained data greatly affects model performance \cite{du2022glam, wei2015submodularity}, though determining the optimal recipe for pretraining data is challenging. Recent research exploring optimization of pretraining data for LLMs at scale includes DoReMi, which trains a small proxy model to produce domain weights for downstream tasks, and then uses the model to resample the dataset for training huge LLMs \cite{xie2024doremi}. However, the question of how to choose data for sample-efficient training of small language models, such as in cases where computational resources are limited, has received little attention.

Psycholinguistic precedent exists for sample-efficient pretraining; children see much less words than a modern LLM yet perform exceptionally well on reasoning tasks. For example, Chinchilla sees over 10000 times the number of words a 13 year old child has ever encountered \cite{choshen2024call}. By the time typical English-speaking children at around 6 years old have obtained adult-level grammatical knowledge  \cite{kemp2005young}, they have seen only around 10-50M words \cite{hart1997meaningful, huebner-etal-2021-babyberta}. In comparison, Llama-3 is trained on 15T tokens \cite{dubey2024llama}. Given the great disparity between the amount of training data an LLM requires and what children require, it seems worthwhile to investigate whether training LLMs can be as sample efficient.

 BabyBERTa \cite{huebner-etal-2021-babyberta} attempts to address this, showing that when training a model on data similar to what is seen by children between the ages 1 and 6, it is able to acquire grammatical knowledge similar to pretrained RoBERTa-base, but with around 15X fewer parameters and 6,000X fewer words; this indicates that utilizing child-directed input may be advantageous for more sample efficient pretraining \cite{huebner-etal-2021-babyberta}. Similarly, \citet{eldan2023tinystories} follow suit, releasing TinyStories, a synthetic dataset of short stories that only contain words that typical 3- to 4-year-old children understand. They demonstrate that TinyStories can be leveraged to train language models with much less parameters than SOTA models, yet still produce coherent output with almost perfect grammar as well as emergent reasoning abilities. Along the same vein, GPT-wee \cite{bunzeck2023gpt} shows that child-directed speech can be used with curriculum learning for simulating children's learning as a potential solution to sample-constrained training.
 
In this paper, we evaluate the effect of different datasets on model performance for sample efficient model training. In our case, we limit our training dataset to 10M words, in accordance with the BabyLM Challenge's super-strict track \cite{choshen2024call}. We consider several different types of datasets, namely child-directed speech (CHILDES), classic books (Gutenberg), a mixed dataset (Mix) and the TinyStories dataset. Experimental results show that smaller models benefit from training on diverse datasets like Mix on the BabyLM evaluation suite \cite{choshen2024call}, but larger models perform better when trained on more complex and rich datasets like Gutenberg. Our findings suggest that the optimal dataset depends on the model size and that neither child-directed speech nor child-directed stories are optimal for language models of any sizes.

\section{Dataset}

For our experiments, we obtained datasets from the BabyLM Challenge \cite{choshen2024call}. Individual categories of 10M-word datasets were procured by extracting the first 10M words from that category in the 100M-word dataset of the BabyLM challenge. We also used Mix, the 10M-word developmentally-plausible corpus of BabyLM, and TinyStories. To measure for complexity in the language of these datasets, we use several readability metrics, including the Flesch reading ease (FRE)
score \cite{Flesch1948ANR}, ARI (Automated Readability Index) \cite{smith1967automated}, and the Gunning fog index \cite{doi:10.1177/002194366900600202}.

For a document $d_i \in \mathcal{C}$, its FRE score is computed as:

\[ \textstyle \mathrm{FRE}(d_i) = 206.835 - (1.015 \cdot \mathrm{ASL}) - (84.6 \cdot \mathrm{ASW}) \]

where ASL is the average sentence length (the number of words divided by the number of sentences) and ASW is the average number of syllables per word (the number of syllables divided by the number of words). Higher FRE scores correspond to simpler texts (e.g., children's literature), while lower scores indicate more complex writing (e.g., machine learning papers). The ARI score is calculated as:

\[ \textstyle \mathrm{ARI}(d_i) = 4.71 \cdot \left(\frac{\mathrm{characters}}{\mathrm{words}}\right) + 0.5 \cdot \left(\frac{\mathrm{words}}{\mathrm{sentences}}\right) - 21.43 \]

Higher ARI scores indicate more complex text requiring higher grade levels to comprehend.
The Gunning fog index score is calculated as:

\[ \textstyle \mathrm{Fog}(d_i) = 0.4 \cdot \left[\left(\frac{\mathrm{words}}{\mathrm{sentences}}\right) + 100 \cdot \left(\frac{\mathrm{complex\ words}}{\mathrm{words}}\right)\right] \]

Like ARI, higher Gunning fog scores indicate more complex text.

Our individual datasets comprise: 
\begin{itemize}
    \item \textbf{CHILDES}: The CHILDES dataset is composed of examples of the human language acquisition process starting from a very young age \cite{macwhinney2000childes}. We constructed a 10 million word training corpus from the CHILDES portion of the small track (100M). We took the first 10M words from the CHILDES portion.

    \item \textbf{Gutenberg}: The Gutenberg dataset is a large dataset composed of English language books \cite{gerlach2018standardizedprojectgutenbergcorpus}. We took the first 10M words from the Gutenberg portion of the small track dataset. 

    \item \textbf{Mix (Default)}: 
    This was the default 10M  dataset for the strict-small track. The split of is displayed below:

\begin{figure}[h]
\centering
\begin{tikzpicture}
\pie[
    radius=2,  
    sum=auto,
    rotate=180,
    text=legend,
    color={blue!60, red!60, yellow!60, green!60, orange!60, purple!60},
]
{
    28.78/CHILDES,
    26.11/Gutenberg,
    20.67/Open Subtitles,
    13.64/Simple Wiki,
    9.34/BNC Spoken,
    1.47/Switchboard
}
\end{tikzpicture}
\caption{\small Default dataset composition}  
\label{fig:dataset-composition}
\end{figure}
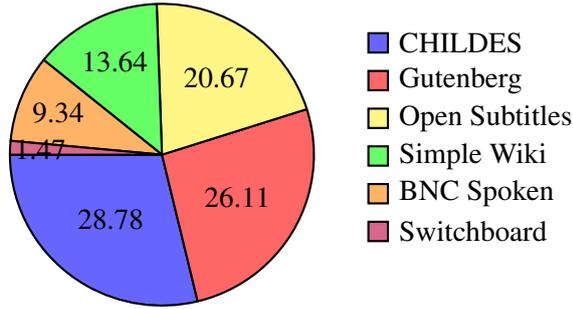

    \item \textbf{TinyStories}: We took the first 10M words from the TinyStories dataset on Hugging Face\footnote{\url{https://huggingface.co/datasets/roneneldan/TinyStories}} \cite{eldan2023tinystories}. (FRE = 105.19)
\end{itemize}

\begin{table}[h]
\centering
\begin{tabular}{lccc}
\toprule
\textbf{Dataset} & \textbf{FRE} & \textbf{Gunning Fog} & \textbf{ARI} \\
\midrule
Mix & 105.89 & 5.62 & 1.59 \\
CHILDES & 115.70 & 2.84 & 0.20 \\
Gutenberg & 87.49 & 9.89 & 7.12 \\
TinyStories & 105.19 & 4.83 & 0.85 \\
\bottomrule
\end{tabular}
\caption{Readability metrics across different datasets. Lower FRE and higher Gunning Fog and ARI scores indicate a more complex dataset.}
\label{table:readability-metrics}
\end{table}

\section{Methodology}
\subsection{Preprocessing}
For both pre-processing and model training, we built off the BabyLlama repository\footnote{\url{https://github.com/timinar/BabyLlama}} \cite{timiryasov2023babyllamaknowledgedistillation}. Following their pre-processing steps, we applied regex-based cleaning and trained a Byte-Pair Encoding tokenizer on the training sets of whatever dataset we were working with. The train and dev sets were split into 128-token chunks, with the model being presented a new random permutation of these chunks in each epoch. Validation loss is computed at the end of each epoch using a fixed, randomly sampled subset of the dev set.

\subsection{Training}
Given that this builds upon the TinyStories paper, focused on dataset optimization for very small language models, we focused mainly on GPT models of sizes 18M, 44M and 97M, which we trained on various datasets. We used this to explore whether different model sizes would affect which dataset performed the best. We trained for 4 epochs, using consistent hyper-parameters. 
Subsequently, we trained a Llama-20M model to confirm that the same pattern regarding dataset complexity is observed in Llama models as well. Lastly, large model baselines of GPT2-705M and Llama-360M are used, as these were the original parent model sizes originally used by last year's BabyLM winning model \cite{timiryasov2023babyllamaknowledgedistillation}.

\subsection{Evaluation}
Evaluation of model performance was done using the BabyLM evaluation suite \cite{choshen2024call}. This consists of the following benchmarks: 
\begin{itemize}
    \item \textbf{BLiMP:} BLiMP (Benchmark of Linguistic Minimal Pairs for English) evaluates language models on their ability to identify grammatical acceptability. It presents pairs of sentences that differ by one linguistic element, testing the model's understanding of 12 areas of English morphology, syntax, and semantics, such as anaphor agreement and filler-gap constructions. It measures how well models assign higher probability to the grammatically correct sentence in each pair. \cite{warstadt2023blimpbenchmarklinguisticminimal}

    \item \textbf{EWoK:} EWoK (Elements of World Knowledge) evaluates language models on their ability to build and apply internal world models. It tests models' understanding of concepts and contexts by presenting them with minimal pairs of scenarios where the models determine the plausibility of context and target combinations. \cite{ivanova2024elementsworldknowledgeewok} 
    
    \item \textbf{GLUE:} GLUE (General Language Understanding Evaluation) evaluates language models on a variety of natural language understanding tasks. It covers tasks such as sentiment analysis, text similarity, question answering, and textual entailment. \cite{wang2019gluemultitaskbenchmarkanalysis} Unlike in the BabyLM evaluation suite, however, we do not do finetuning in this case and run it as a zero-shot evaluation due to computational constraints. 
\end{itemize}

\section{Results and Discussion}
\begin{table*}[t]
\centering
\resizebox{0.95\textwidth}{!}{%
\begin{tabular}{lccccc} 
\toprule
\textbf{Model} & \textbf{Dataset} & \textbf{BLiMP Supplement} & \textbf{BLiMP Filtered} & \textbf{EWoK} & \textbf{Macroaverage} \\ 
\midrule
GPT2-18M & CHILDES & 52.8 & 58.2 & 50.5 & 53.83 \\
         & Gutenberg & 55.7 & 62.4 & 50.3 & 56.13 \\
         & Mix & \textbf{55.9} & \textbf{63.7} & 49.7 & \textbf{56.43} \\
         & TinyStories & 55.2 & 57.5 & \textbf{50.7} & 54.47 \\ 
\midrule
GPT2-44M & CHILDES & 55.3 & 57.8 & \textbf{51.2} & 54.77 \\
         & Gutenberg & 57.6 & 63.0 & 50.0 & 56.87 \\
         & Mix & \textbf{58.2} & \textbf{65.6} & 50.4 & \textbf{58.07} \\
         & TinyStories & 52.8 & 57.1 & 50.4 & 53.43 \\ 
\midrule
GPT2-97M & CHILDES & 49.7 & 60.5 & 49.6 & 53.27 \\
         & Gutenberg & \textbf{59.0} & \textbf{65.3} & \textbf{51.1} & \textbf{58.47} \\
         & Mix & 58.0 & 66.0 & 50.6 & 58.20 \\
         & TinyStories & 54.6 & 59.1 & 50.3 & 54.67 \\ 
\midrule
Llama-20M & CHILDES & 53.4 & 57.9 & 50.2 & 53.83 \\
          & Gutenberg & \textbf{57.4} & 60.0 & \textbf{50.6} & \textbf{56.00} \\
          & Mix & 56.6 & \textbf{62.8} & 50.2 & 56.53 \\
          & TinyStories & 46.7 & 51.1 & 49.8 & 49.20 \\ 
\hline
\hline
GPT2-705M & Gutenberg & \textbf{59.9} & \textbf{66.8} & \textbf{50.6} & \textbf{59.10} \\
          & Mix & 56.7 & 66.1 & 50.6 & 57.80 \\ 
\midrule
Llama-360M & Gutenberg & \textbf{56.7} & \textbf{66.5} & 50.2 & \textbf{57.80} \\
           & Mix & 56.6 & 62.8 & \textbf{50.5} & 56.63 \\
\bottomrule
\end{tabular}
}
\caption{Summary of BLiMP filtered, BLiMP supplement, EWoK results, and Macroaverage for various models and datasets}
\label{table:results}
\end{table*}

Overall, our results demonstrate that the effectiveness of a training dataset is dependent on the model size. Specifically, smaller models (with fewer parameters) benefit more from training on a diverse dataset like Mix, while larger models show improved performance when trained on the Gutenberg dataset. As shown in Table 2, for smaller models like GPT2-18M and GPT2-44M, Mix consistently achieves the best performance on BLiMP, scoring 63.7 and 65.6 respectively on BLiMP Filtered, and 55.9 and 58.2 on BLiMP Supplement. However, as we move to larger models like GPT2-97M and GPT2-705M, the Gutenberg dataset takes the lead, achieving the highest scores across most metrics (59.0 and 59.9 on BLiMP Supplement, 65.3 and 66.8 on BLiMP Filtered). We see this also extend to the Llama models as well, where the larger Llama-360M performs best with Gutenberg data (56.7 on BLiMP Supplement and 66.5 on BLiMP Filtered), while the smaller Llama-20M shows mixed results between Gutenberg and Mix. Interesting, both CHILDES and TinyStories consistently underperform across all model sizes, with scores typically lower than both Mix and Gutenberg datasets.
On the other hand, we see a very different story when looking at macro average GLUE scores for the models (Table 3), with TinyStories performing well for small models and CHILDES performing well for the big model. However, when examining the GLUE subtasks further, we do not see a clear trend on which dataset type results a stronger performance, and cannot conclude a clear trend here. 

\begin{table*}[t]
\centering
\resizebox{\textwidth}{!}{%
\begin{tabular}{llccccccccccc} 
\toprule
\textbf{Model} & \textbf{Dataset} & \textbf{MRPC} & \textbf{MultiRC} & \textbf{QNLI} & \textbf{SST-2} & \textbf{BoolQ} & \textbf{MNLI} & \textbf{QQP} & \textbf{WSC} & \textbf{RTE} & \textbf{Cola (MCC)} & \textbf{Macro Average} \\ 
\midrule
GPT2-18M & CHILDES     & 34.31 & 45.50 & 50.92 & 53.67 & 58.59 & 32.42 & 42.47 & 38.46 & 48.20 & -0.07 & 40.45 \\
         & Gutenberg   & 35.78 & 52.56 & 49.52 & 47.94 & 46.73 & 32.74 & 60.68 & 61.54 & 53.24 & 0.05 & 44.08 \\
         & Mix         & 58.33 & 44.35 & 47.14 & 47.71 & 57.00 & 32.42 & 46.77 & 46.15 & 44.60 & 0.03 & 42.45 \\
         & TinyStories & 60.78 & 42.86 & 51.72 & 51.83 & 62.63 & 32.56 & 50.54 & 42.31 & 48.20 & 0.06 & 44.35 \\ 
\midrule
GPT2-44M & CHILDES     & 46.57 & 42.41 & 51.13 & 47.71 & 55.96 & 32.84 & 41.52 & 53.85 & 46.76 & 0.07 & 41.88 \\
         & Gutenberg   & 64.71 & 45.54 & 50.88 & 50.92 & 60.98 & 31.85 & 37.32 & 38.46 & 43.88 & -0.02 & 42.45 \\
         & Mix         & 52.94 & 47.07 & 50.62 & 48.17 & 55.23 & 32.42 & 54.06 & 38.46 & 42.45 & 0.03 & 42.14 \\
         & TinyStories & 45.59 & 53.09 & 47.04 & 48.39 & 42.26 & 33.19 & 62.01 & 59.62 & 54.68 & -0.06 & 44.58 \\ 
\midrule
GPT2-97M & CHILDES     & 57.35 & 53.42 & 49.27 & 50.23 & 44.59 & 35.76 & 62.47 & 61.54 & 53.96 & 0.06 & 46.86 \\
         & Gutenberg   & 54.90 & 47.69 & 50.62 & 53.21 & 54.98 & 31.46 & 38.67 & 38.46 & 43.17 & 0.03 & 41.32 \\
         & Mix         & 47.05 & 49.88 & 48.57 & 50.00 & 44.10 & 33.48 & 61.82 & 61.54 & 56.12 & -0.05 & 45.25 \\
         & TinyStories & 65.20 & 43.61 & 50.40 & 51.83 & 62.08 & 32.03 & 38.05 & 44.23 & 50.36 & 0.07 & 43.79 \\
\bottomrule
\end{tabular}
}
\caption{Detailed GLUE scores for various GPT model sizes and datasets}
\label{table:glue_results_detailed}
\end{table*}

 \subsection{Dataset and model performance}

Model performance results on various datasets was observed in table 2. 
Small models, such as GPT2-18M and GPT2-44M, have limited capacity due to fewer parameters. This constraint affects their ability to capture complex linguistic patterns and nuanced language structures. Datasets like Gutenberg with a relatively lower FRE score (87.49) contain wider vocabulary, more intricate syntax, and nuanced semantic meaning. Due to their limited capacity, small models cannot fully learn from the complexity of the dataset. They oversimplify the language patterns, leading to high bias and poor generalization. This underfitting results in lower performance on evaluation benchmarks. 

In contrast, larger models, such as GPT2-97M, GPT2-705M, and LLaMA-360M, possess greater capacity to learn and represent complex patterns due to their increased number of parameters. Because the Gutenberg dataset, consisting of a diversity of subject materials \cite{gerlach2018standardizedprojectgutenbergcorpus}, offers the most nuanced sentence structures and vocabulary out of all the datasets, it could be argued that diversity within the dataset may be more important than having a diverse basket of datasets for models with a higher number of parameters.

\subsection{Dataset Convergence}
In our experiments, CHILDES converged faster than either then Gutenberg or the Mix datasets for both GPT2-44M and GPT2-18M models. This can be observed in figure 2 and 3 below, and can be explained by the nature of CHILDES dataset. The higher FRE score (115.70) of this child-directed speech dataset indicates simpler grammatical structures, shorter sentences, and straightforward syntax compared to the adult-oriented language found in datasets like Gutenberg or Mix. In addition, because caregivers frequently repeat words and phrases when interacting with children, the dataset is characterized by high repetition, making the learning task of capturing the underlying structures and relationships in the data easier and faster to converge quickly during training. In short, due to the low perplexity of the CHILDES dataset, the model has less uncertainty in predicting the next word in a sequence, resulting in a smoother loss landscape and simplifying the learning task.

\begin{figure}[h]
    \centering
    \includegraphics[width=1.03\linewidth]{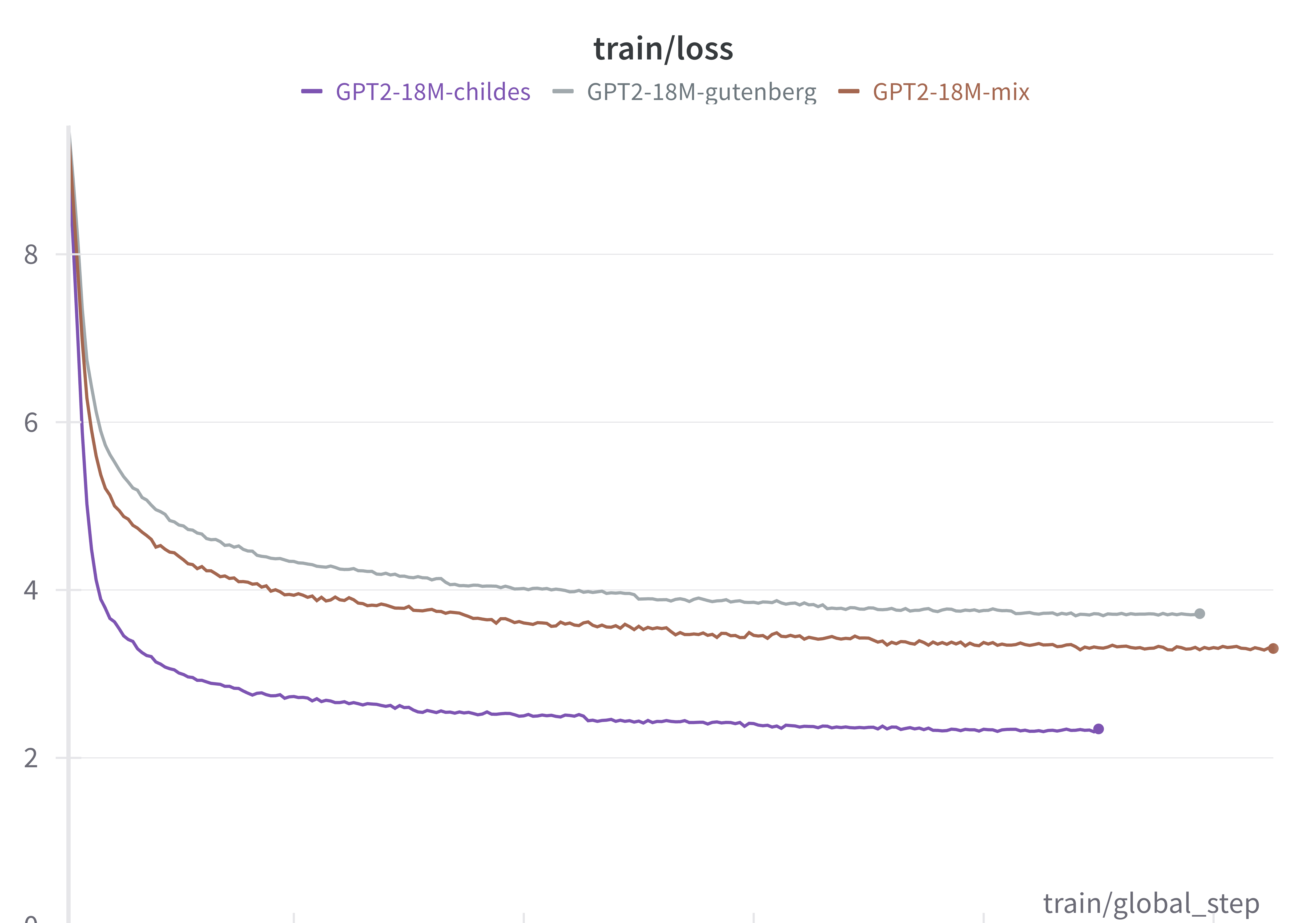}
    \caption{Train loss when training GPT2-18M on various datasets}
    \label{fig:enter-label}
\end{figure}

\begin{figure}[h]
    \includegraphics[width=1.03\linewidth]{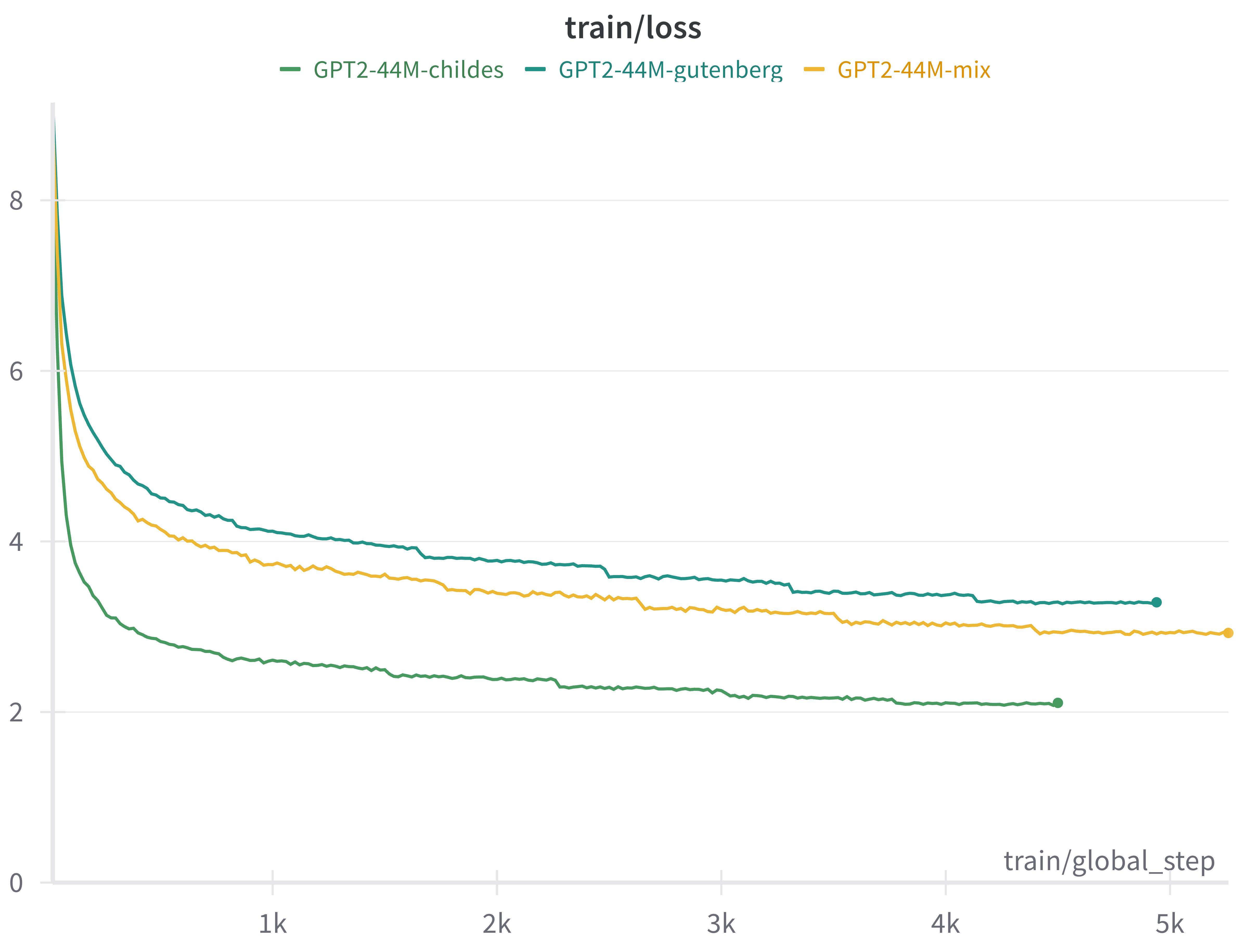}
    \caption{Train loss when training GPT2-44M on various datasets}
    \label{fig:enter-label}
\end{figure}

\subsection{Underperformance of Child-directed and Synthetic Datasets}
Neither the CHILDES nor TinyStories datasets performed very well on the BLiMP or EWoK evaluation suite \cite{choshen2024call}. The CHILDES dataset consistently underperformed no matter the model size, suggesting that child-directed speech may not be not advantageous for training a robust model. This is consistent with the lack of success in implementing curriculum learning for child data in the previous BabyLM challenge \cite{bunzeck2023gpt}. In their paper, Bunzeck and Zarrieß noted that the integration of more sophisticated linguistic factors into the training process might be needed, as their curriculum approach based on prototypicality measures didn't effectively capture the language acquisition process they were looking for. 

Considering the strong performance of TinyStories in \cite{eldan2023tinystories}, and the fact that we adopted the same GPT-44M architecture as in paper, with a hidden size of 768, 2 layers and 8 heads, we were surprised by the poor performance of the TinyStories dataset. That said, we only used a 10M subset of TinyStories, and given its limited vocabulary and grammatical range (and higher FRE score of 105.19), perhaps there was insufficient diversity and exposure to new formats as previously discussed. Additionally, we utilized different benchmarks. The BLiMP and EWoK benchmarks assess a model's understanding of complex grammatical rules and world knowledge; this is not likely to be adequately covered by the TinyStories dataset. In short, models trained on TinyStories may lack exposure to the types of linguistic phenomena these benchmarks evaluate. 

The disparity in TinyStories’ performance across benchmarks likely stems from the divergent linguistic and cognitive demands of each dataset. GLUE evaluates general-purpose natural language understanding (NLU) tasks, such as sentiment analysis and paraphrase identification, which align well with the broad, semantic patterns learned from narrative content in TinyStories. In contrast, BLiMP emphasizes fine-grained syntactic and grammatical competence, while EWoK assesses factual reasoning and contextual world knowledge—skills that TinyStories’ simplified narrative structure and limited syntactic diversity do not comprehensively support. Consequently, while TinyStories provides effective training for NLU, it lacks the complexity required for the precise linguistic and knowledge-based reasoning assessed by BLiMP and EWoK.

On the whole, however, we do not see the huge performance gains that were reported in the original TinyStories paper. The success of TinyStories in the original paper may perhaps be partially attributed to the narrative structure of the data, which provides contextual coherence and sequential dependencies that models can leverage. However, given that the Gutenberg dataset also contains narrative texts but with more complicated language and storylines, it offers better training data for models to learn general language patterns.

\section{Limitations}
Our study has several limitations. First, we used consistent hyper-parameters across all experiments for comparability, but this may not have been optimal for each model-dataset pair. Tuning hyper-parameters individually could have yielded better performance. 

Second, the BLiMP and EWoK benchmark assess linguistic competence on tasks on represented in datasets such as TinyStories or CHILDES, potentially biasing the evaluation. In short, there is a mismatch between the training data afforded by child datasets and the test set.

Lastly, due to computational limitations, models were trained for only four epochs. Longer training might have allowed models to better capture the nuances of the datasets.

\section{Conclusion and Future Work}
In this paper, we investigated the impact of dataset composition on the performance of small language models in a sample-efficient training regime. By training models of varying sizes on different datasets limited to 10 million words, we sought to identify which types of data are most beneficial for language acquisition in resource-constrained settings. 

We found that tiny models (e.g., GPT2-18M and GPT2-44M) performed best when trained on the Mix dataset, which offers a diverse combination of language inputs, while slightly larger small language models achieved superior performance when trained on the Gutenberg dataset, leveraging its linguistic richness. In contrast, models trained on CHILDES or TinyStories underperformed regardless of size. 

For future work, a more thorough investigation of other types of data sources such as news articles, scientific texts, and conversational data might better tease out the optimal dataset for model performance. Additionally, it might be useful to explore curriculum learning, which presumable models the developmental process of a language learning child. 

Widening the benchmarks beyond GLUE and BLiMP tasks to coherent text generation, as well as scaling dataset sizes and tasks would allow for a more comprehensive and robust study as well. 

\section*{Acknowledgments}

We thank Stanford University for their support for this paper. 

\bibliography{custom}


\end{document}